\documentclass{article}
\usepackage{ijcai26}

\usepackage{times}
\usepackage{soul}
\usepackage{url}
\usepackage[hidelinks]{hyperref}
\usepackage[utf8]{inputenc}
\usepackage[small]{caption}
\usepackage{graphicx}
\usepackage{amsmath}
\usepackage{amsthm}
\usepackage{booktabs}
\usepackage{algorithm}
\usepackage[switch]{lineno}

\newcommand{\citep}[1]{\cite{#1}}
\newcommand{\citet}[1]{\citeauthor{#1} [\citeyear{#1}]}

\usepackage{placeins}
\usepackage{enumitem}
\usepackage{relsize}

\usepackage[textsize=tiny]{todonotes}
\usepackage{amsmath}
\usepackage{graphicx}
\usepackage{amsfonts}       
\usepackage{nicefrac}       
\usepackage{microtype}      
\usepackage[dvipsnames]{xcolor}         
\usepackage[noEnd=trueb]{algpseudocodex}
\usepackage{subcaption}
\usepackage{xcolor}
\theoremstyle{plain}
\usepackage{siunitx}
\usepackage{wrapfig}
\usepackage{bbm}
\DeclareMathOperator*{\argmin}{arg\,min}
\DeclareMathOperator*{\argmax}{arg\,max}

\newtheorem{property}{Property}

\title{Scalable Decision-Focused Learning through Cost-Sensitive Regression} 

\author{
Noah Schutte$^1$\and
Senne Berden$^2$\and
Tias Guns$^2$\and
Krzysztof Postek$^3$\and
Neil Yorke-Smith$^1$
\\
\affiliations
$^1$Delft University of Technology\\
$^2$KU Leuven\\
$^3$Independent Researcher
}

\begin{document}

\maketitle

\begin{abstract}
Many real-world combinatorial 
problems involve uncertain parameters, which can be predicted given contextual features and historical data. These `predict-then-optimize' or `contextual optimization' problems have gained significant attention: 
end-to-end training methods can now minimize the downstream task cost rather than the predictive error.
However, despite their effectiveness, these decision-focused learning (DFL) approaches often rely on repeated solving of the underlying combinatorial optimization problem during training, making them computationally expensive and difficult to scale.
We reframe
the learning problem as a \emph{cost-sensitive multi-output regression problem}: multi-output due to the combinatorial problem having multiple uncertain parameters, and cost-sensitive due to the downstream task cost being the real target. Our technical contribution is the formalization of multiple loss function components that follow from this reframing: cost-insensitive normalization, decision-aware asymmetric penalization of over- and underpredictions, and instance-based costs that mimic the true downstream task-based loss locally. These components require zero or one solve per training data instance, while requiring no further solves during training. Experiments show that the combination of loss components achieves comparable downstream task quality to the state of the art, while being significantly more efficient, enabling scaling to problem sizes that have not been tackled before with DFL.

\end{abstract}



\section{Introduction}
\label{sec:intro}

Prediction problems often relate to some downstream decision-making task. For example, short-term electricity demands are predicted based on historical patterns in order to decide when to buy or sell energy. Many of these problems involve decision-making over multiple predictions at once, for example, predicting and deciding ahead for the coming 24 hours, or choosing among sets of goods subject to capacity limits.

Such predict-then-optimize problems can be framed as \emph{contextual stochastic optimization problems}~\citep{sadana2025}: contextual due to features and historical data being present, and stochastic due to the uncertainty of the values that are to be predicted. To train the predictive model, the recent 
field of decision-focused learning 
\citep{spo,dfl_survey} has developed novel end-to-end learning techniques that can back-propagate through combinatorial problems. This is challenging from a 
learning perspective because combinatorial optimization problems incur zero gradients almost everywhere, as 
optimal solutions 
generally remain unchanged under small perturbations of 
the
input parameters.

Despite the effectiveness of these DFL techniques, they generally require solving the optimization problem in each forward pass in the training procedure, and are therefore computationally expensive. To alleviate this computational cost, several approaches have been proposed: using problem relaxations \citep{melding}, caching optimal decisions for reuse later \citep{contrastive}, or adopting surrogate losses that require solving cheaper quadratic programs \citep{tang2024cave} or no solves at all \citep{berden2025}.

This paper furthers this last category by taking a cost-sensitive perspective: we reframe the predict-then-optimize problem as a \emph{cost-sensitive multi-output regression problem}.  While cost-sensitive regression has already been used for simple decision-making problems, such as the buy-or-sell energy example, where overpredictions can be less costly than underpredictions, this does not readily translate to combinatorial problems.  To do so, we adjust standard accuracy-based losses by introducing multiple cost-sensitive loss components with respect to the optimization problem, without requiring solve calls during training and while preserving differentiability.

Our contributions can be summarized as follows:
\begin{itemize}[leftmargin=*,nosep]
    \item 
    Formalizing cost-sensitive multi-output regression as a framework for DFL, where the loss function is asymmetric and multiplicative costs are defined at two levels: over multiple predictor outputs and over multiple (all) instances.
    \item 
    Multiple loss components within this framework that keep the loss differentiable, of which two do not require any solves and one requires a single solve call per instance. 
    \item 
    Experiments showing that: the loss components are empirically monotonic; their combination leads to similar decision quality compared to SOTA surrogate loss functions that require solver calls during training; and with them we can scale to problem sizes that have not been considered before in a DFL setting, showing significant improvement in training time and/or decision quality.
\end{itemize}

\section{Problem Formulation}
\label{sec:problem}

\subsection{Decision-Focused Learning}
Observing some contextual information in the form of feature values $z \in \mathbb{R}^d$, coming from some probability distribution $z \sim \mathcal{Z}$, the goal
of the decision maker is to solve the following stochastic
optimization problem with linear objective and bounded, non-empty feasibility set $X$ for $c \in \mathbb{R}^d$:
\begin{align} \label{eq:sto}
    \max_{x \in X} \mathbb{E}_{c \sim \mathcal{C}_z}[c^T x|z].
\end{align}

Uncertain parameters $c = [c_1\; c_2\; \ldots\; c_d]$, where $c_j \in \mathbb{R}$, are unknown but correlated with a known feature vector $z \in \mathbb{R}^k$ according to some distribution $\mathcal{C}_z$. Although $\mathcal{Z}$ and $\mathcal{C}_z$ are not known, we assume access to a data set of instances $D = \{(z^{(i)}, c^{(i)}, x^\star(c^{(i)}))\}_{i=1}^n$, where $x^\star(c) = \argmax_{x \in X}  c^Tx$. Data set $D$ is used to train a model $\phi_\theta$ with learnable parameters $\theta$ -- usually a linear regression model or a neural network -- which makes predictions $\hat{c}$.

Unlike conventional regression, the objective in training is \textit{not} to maximize the accuracy of predicted costs $\hat{c}$. Rather, the aim is to make predictions that maximize the quality of the resulting decisions. This is measured by the \textit{regret}, which expresses the suboptimality of the made decisions $x^\star(\hat{c})$ with respect to true costs $c$ (lower is better).
\begin{equation*}
    \ell_\text{regret}(\hat{c}, c) = c^T x^\star(c) - c^T x^\star(\hat{c})
\end{equation*}

Hence the main objective becomes to find:
\vspace{-1mm}
\begin{equation} \label{eq:totalregret}
 \argmin_\theta \sum_{i=1}^n \ell_\text{regret}(\phi_\theta(z^{(i)}), c^{(i)})
\end{equation}
In general, gradient descent is effective for regression problems. However, significant challenges arise when using regret as a loss function. Specifically, when feasibility set $X$ is discrete or polyhedral (i.e., $X = \{x \in \mathbb{R}^d: Ax \leq b, A \in \mathbb{R}^{m \times d}, b \in \mathbb{R}^m\}$), the gradient of the regret is zero almost everywhere. In the discrete case, the mapping from parameters $\hat{c}$ to optimal decisions $x^\star(\hat{c})$ maps a continuous space to a discrete one and is therefore piecewise constant, resulting in zero gradients almost everywhere. In the polyhedral case, optimal solutions lie at vertices of the feasible region, inducing a similar continuous-to-discrete mapping and again leading to zero-valued gradients.
Because of this, a multitude of smoothing techniques and surrogate loss functions have been proposed \citep{dfl_survey}. In this work, we tackle the underlying problem by defining and applying a cost-sensitive learning framework to obtain a cost-sensitive differentiable surrogate loss function.

\subsection{Cost-Sensitive Learning}

Cost-sensitive learning aims at minimizing costs associated with predictions, instead of minimizing the predictive error \citep{granger1969,elkan2001}. In the classification domain, where most work in cost-sensitive learning is done, misclassifications are assigned different costs. For example, type II errors (i.e., false negatives) are often more costly than type I errors (i.e., false positives): classifying a catastrophic event as not-happening while it is in fact happening is generally more costly than the other way around. Additionally, instance-dependent costs can be considered \citep{vanderschueren2022}. For example: data is recorded on a daily basis, and the costs of catastrophic event mispredictions vary by day. To formalize these costs, we first introduce regular classification.

Given a data set $D = \{(z^{(i)}, y^{(i)})\}_{i=1}^n$, with features $z \in \mathbb{R}^k$ and labels $y \in \{1, ...,d\}$, regular classification would aim to find classifier $h_\theta : \mathbb{R}^k \to \{1, ..., d\}$ such that the number of misclassifications is minimized, i.e.:
\vspace{-1mm}
\begin{equation*}
    \argmin_\theta \sum_{i=1}^n \mathbbm{1}_{h_\theta(z^{(i)}) \neq y^{(i)}}
\end{equation*}
where $\mathbbm{1}$ denotes an indicator function. In cost-sensitive classification, one defines cost vectors $C^{(i)} \in \mathbb{R}^d$ for each instance $i \in \{1, \dots, n\}$, where a cost is defined for each potential classification for each instance. This 
leads to
classifier $h_\theta$ by: 
\vspace{-1mm}
\begin{equation*}
    \argmin_\theta \sum_{i=1}^n C^{(i)}_{h_\theta(z^{(i)})} 
\end{equation*}


Due to the discrete nature of classification, we observe discrete incurred costs $C_j^{(i)}$. However, this is different in regression, where the output of the predictive model is continuous and therefore requires a cost-\emph{function} to be specified.
Accordingly, in cost-sensitive regression, these cost-functions are studied \citep{zhao2011}, i.e., we observe $D = \{(z^{(i)}, c^{(i)})\}_{i=1}^n$, with features $z \in \mathbb{R}^k$ and (single-output) predictions $c \in \mathbb{R}$. For example, the task is not to predict whether a catastrophic event will occur, but to predict its magnitude. In this scenario, it is (similarly) often better to overpredict and be well prepared than to underpredict. This leads to the use of asymmetric loss functions, which have been studied extensively in the field of statistics \citep{zellner1986}, and were more recently applied in cost-sensitive learning \citep{bansal2008}. One commonly used asymmetric loss is the pinball loss, used for quantile regression. The pinball loss aims at finding predictions such that a certain quantile is achieved, i.e., $\tau$ percent of predictions are overpredictions and $1-\tau$ predictions are underpredictions, where $\tau \in [0,1]$.  Formally, $\ell_\tau: \mathbb{R}^2 \to \mathbb{R}_+$, such that:
\vspace{-1mm}
\begin{equation} 
\ell_{\tau}(\hat{c}, c) = \begin{cases} 
\tau e(\hat{c}, c) & \text{if }  \hat{c} \leq c \\
(1 - \tau) e(\hat{c}, c) & \text{if } \hat{c} > c
\end{cases}
\end{equation}
where $\hat{c}=\phi_\theta(z)$ is the prediction obtained from predictive model $\phi_\theta: \mathbb{R}^k \to \mathbb{R}$, and $e(\hat{c},c)=|\hat{c}-c|$ is the absolute error, but can also be defined as the squared error $e(\hat{c},c)=(\hat{c}-c)^2$, resulting in the squared pinball loss, or asymmetric mean squared error. Based on these definitions of cost-sensitive classification and regression, we will formalize cost-sensitive multi-output regression and present it as a framework for DFL.

\section{Cost-Sensitive Decision-Focused Framework}
\label{sec:method}

We reframe the DFL setting as a \textit{cost-sensitive multi-output regression} problem. In this framework we perform multi-output regression with predictive model $\phi_\theta: \mathbb{R}^k \to \mathbb{R}^d$. Due to the regression problem being multi-output, we allow for costs per instance and costs per output. Additionally, we allow for the use of an asymmetric loss function. Given some base error function $e: \mathbb{R}^{2} \to \mathbb{R}$,
we can generically define the cost-sensitive total multi-output regression loss as:
\begin{equation}
\label{eq:cost_sensitive_multi_output_regression_loss}
    \sum_{i=1}^n C^{(i)} \frac{1}{d} \sum_{j=1}^d  (\tau_j^{(i)} \mathbbm{1}_{\hat{c}^{(i)}_j \leq c^{(i)}_j} + (1-\tau_j^{(i)}) \mathbbm{1}_{\hat{c}^{(i)}_j > c^{(i)}_j}) \hspace{1mm} e(\hat{c}^{(i)}_j, c^{(i)}_j),
\end{equation}
where costs $C^{(i)}$ are instance-dependent and asymmetric loss ratios $\tau_j^{(i)}$ are also output-dependent. We use the mean as aggregation over the output-dimension. Recall that, in the DFL setting, the actual objective is to minimize the regret. To do this using the cost-sensitive multi-output regression framework, we will present: 1) instance-based costs that approximate regret locally, 2) an asymmetric one-sided loss that is decision-aware and 3) a scale-invariant transformation to the parameters (normalization), to which the regret is cost-insensitive. 

Note that one can recover the mean squared error (MSE) from the formalization in \eqref{eq:cost_sensitive_multi_output_regression_loss}, by taking the squared difference as error function ($e(\hat{c},c)=(\hat{c} - c)^2$), setting $\tau_j^{(i)} = 0.5$ for all $i, j$, and setting instance-dependent costs $C^{(i)} = 2$ for all $i$. 
\vspace{-1mm}
\begin{equation*}
\ell_\text{MSE}(\hat{c}, c) = \frac{1}{d}\sum_{j=1}^d(\hat{c}_j-c_j)^2.
\end{equation*}

Being often used in regular regression, we consider MSE as a baseline. Despite MSE being decision-unaware, predictions with a low MSE often lead to better decisions than predictions with a high MSE. 
Further, MSE obeys:
\begin{property}[Regret consistent]\label{prop:zero}
Loss $\ell$ is regret consistent if:
\begin{equation*}
    \ell(\hat{c}, c) = 0 \implies \ell_\text{regret}(\hat{c}, c)=0
    \label{prop:zero:eq}
\end{equation*}
\end{property}
\noindent
since $\ell_\text{MSE}(\hat{c}, c) = 0$ only holds when $\hat{c} = c$, and $\ell_\text{regret}(c, c) = 0$ for all $c \in \mathbb{R}^d$. This property is important for our loss, as without it, 
zero loss does not necessarily result in zero regret.


\subsection{Costs-Based}

From a cost-sensitive perspective, not every instance is equal. We aim at defining instance-based costs $C^{(i)}$ for all $i \in \{1, \dots, n\}$, valuing instances with respect to their impact on our target objective of total regret \eqref{eq:totalregret}. We will do this by taking the following steps, similar to \citet{lawless2022}: First, we train with a base loss function $\ell$ to get a baseline predictive model $\bar{\phi}$. After this, we compute instance-based costs $C^{(i)}$ to define a new loss:
$$\ell_i^{(\text{C})}(\hat{c},c) = C^{(i)} \cdot \ell(\hat{c},c).$$ 

We determine $C^{(i)}$ by approximating $\ell_\text{regret}$ locally: After we trained a baseline predictive model $\bar{\phi}$, we are able to determine how our base loss $\ell$ relates to our target loss $\ell_\text{regret}$, at least locally around predictions $\bar{c} = \bar{\phi}(z)$. We do this by computing regret on all instances in our dataset $\ell_\text{regret}(\bar{c}^{(i)}, c^{(i)})$ for $i \in \{1, \dots, n\}$. We divide this value by the realized base loss values $\ell(\bar{c}^{(i)},c^{(i)})$ to obtain this local relationship. This is what we use as instance-based costs: 
\vspace{-1mm}
\begin{equation} 
    C^{(i)} = \frac{\ell_\text{regret}(\bar{c}^{(i)}, c^{(i)})}{\ell(\bar{c}^{(i)},c^{(i)})}. 
\end{equation}
Note that $C^{(i)}$ remains fixed when we train with $\ell^{(C)}_i$. A property of this loss is that at our baseline predictions, the instance-based costs loss equals total regret $\sum_{i=1}^n \ell_i^{(\text{C})}(\bar{c}^{(i)},c^{(i)}) = \sum_{i=1}^n \ell_\text{regret}(\bar{c}^{(i)},c^{(i)})$, which is the main objective we are aiming to minimize. Assuming $\ell$ and $\ell_\text{regret}$ are (locally) linearly correlated, every percentage adjustment in $\ell$ now corresponds to the same percentage adjustment in $\ell_\text{regret}$. This makes loss $\ell^{(\text{C})}_i$ able to redistribute loss values of $\ell$ in a way that is likely to improve regret. This (local) linear correlation is a strong assumption that does not hold in the DFL setting we are targeting, however it is not an unreasonable approximation.
Additionally, $\ell_i^{(\text{C})}$ has Property~\eqref{prop:zero} if base loss $\ell$ has it as well, as long as $C^{(i)} \neq 0 \quad \forall i \in \{ 1, \dots, n\}$. 
\vspace{-1mm}
\begin{proof}
    Take any $C \in \mathbb{R}$ such that $C \neq 0$:
    $$C\cdot\ell(\hat{c},c) = 0 \implies \ell(\hat{c},c) = 0 \implies \ell_\text{regret}(\hat{c},c)=0\qedhere$$
\end{proof}
In general, $C^{(i)}=0$ is a problem. It means that $\ell_i^{(\text{C})}(\hat{c},c)= 0$ for any $\hat{c},c \in \mathbb{R}^d$, effectively ignoring instance $i$ during training. $C^{(i)}=0$  happens when $\ell_\text{regret}(\bar{c}^{(i)},c^{(i)}) = 0$. To adjust for this we set instances that attain zero regret at the baseline prediction equal to the average instance-based cost. Define $N^+=\{i : \ell_\text{regret}(\bar{c}^{(i)}, c^{(i)}) > 0\}$. $C^{(k)} = \bar{C} = \frac{1}{|N^+|}\sum_{i \in N^+}^n C^{(i)}$, for all $k \in \{1, \dots,n\} \setminus N^+$. This average is based on the following principle: For all instances for which we observe a regret of zero with our baseline predictive model, this zero-regret was obtained by having an instance-based cost of 1. One can interpret this as having a weight of $\frac{1}{n}$, relative to the whole dataset. Since zero regret is favourable and we do not have an estimation of the local regret-to-base-loss relationship, we aim to keep this weight equal, achieved by setting the instance-based costs to the average.

Instance-based costs $C^{(i)}$ give higher weights to instances with higher regret given the baseline predictor. This approach has been applied by \citet{lawless2022} by reweighting the prediction loss by the obtained regret from the baseline predictor, including a hyperparameter $w$ that determines how much the costs are weighted. This makes their loss function:
\begin{equation} \label{eq:lawless}
     \ell_{i}^{\text{W}(w)}(\hat{c}, c)= (w C^{(i)} + (1-w)) \cdot \ell(\hat{c}, c), 
\end{equation}
where $C^{(i)} =\ell_\text{regret}(\bar{c}^{(i)}, c^{(i)})$. There are two differences compared to our loss: 1) We set the baseline regret relative to baseline base loss, 2) We do not use a hyperparameter $w$. To validate our design, we run experiments comparing to \eqref{eq:lawless} with different hyperparameter values $w$. 

Iteratively 
updating the instance-based costs would be a natural extension. However we found that this does not improve performance. We refer to the Appendix for discussion.

\subsection{One-Sided} \label{sec:asym}
\begin{figure}[tbp]
    \setlength{\abovecaptionskip}{4pt}
    \setlength{\belowcaptionskip}{-4pt}
    \centering
    \includegraphics[width=\linewidth]{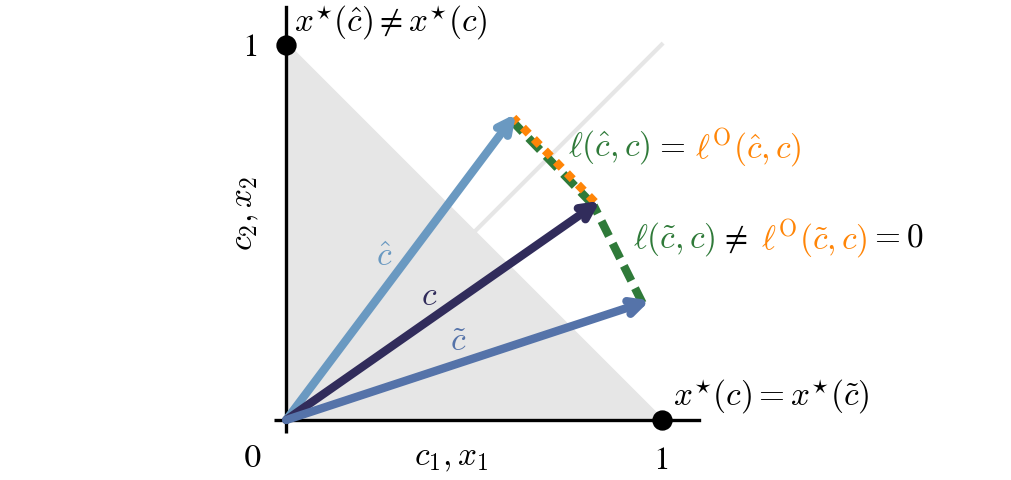}
    \caption{Example of one-sided loss on a linear problem $\max_{x \in X} c^T x$ with feasible space (shaded) $X= \{(x_1,x_2): x_1+x_2 \leq 1, x_1 \geq 0, x_2 \geq 0\}$. Shaded line denotes $c_1=c_2$. $\ell$ = $\ell_\text{MSE}$. $\ell_\text{MSE}(\hat{c},c)$ and $\ell_\text{MSE}(\tilde{c},c)$ have similar values (green dashes); however, $\ell_{\text{regret}}(\hat{c},c) = c_1 - c_2 > 0$ and $\ell_{\text{regret}}(\tilde{c},c) = 0$. In contrast, with the one-sided loss, $\ell_\text{MSE}^{\text{O}}(\tilde{c},c) = \ell_{\text{regret}}(\tilde{c},c) = 0$, because $\tilde{c}$ overpredicts where the optimal decision is 1, and underpredicts where it is $0$. The loss for $\hat{c}$ remains $\ell_\text{MSE}^{\text{O}}(\hat{c},c) = \ell_{\text{MSE}}(\hat{c},c) \gg 0$.}
    \label{fig:asym}
\end{figure}
In general, accuracy metrics like MSE are symmetric in the direction of the error, i.e., overpredictions and underpredictions are valued equally. However, for decision making these do not necessarily have the same impact. Because of this, cost-sensitive regression studies asymmetric loss functions.

If we have access to the optimal decisions of the instances in our training data, we can use this information to see the impact of the direction of the error. This makes the loss decision-aware, which is especially effective for binary problems with linear objective. We refer to the Appendix for the more general case with linear objective and bounded feasible space.

For binary linear problems, the optimal decisions in the training data contain relevant information. 
In MSE, a predicted parameter $\hat{c}_j$ contributes $(c_j - \hat{c}_j)^2$ to the loss. However, an overprediction $\hat{c}_j > c_j$ is not necessarily as bad as an underprediction $\hat{c}_j < c_j$. In our case, one of these directions does not have to be penalized at all, based on the following: If one of the decision variables should be chosen (i.e., $x^\star_j(c) = 1$), we do not mind overprediction and therefore only penalize underpredictions ($\tau_j = 1$). If the decision variables should not be chosen (i.e., $x^\star_j(c) = 0$), then we only penalize overpredictions ($\tau_j = 0$). The situation is reversed in case of a minimization problem. 

This makes this loss not just asymmetric, but one-sided. Given some error function $e: \mathbb{R}^2 \to \mathbb{R}$, the one-sided loss (denoted as component `O') is defined by:
\begin{equation*}
\ell^{(\text{O})}(\hat{c}, c) = \frac{1}{d}\sum_{j=1}^{d} \mathbbm{1}(\hat{c}_j,c) \cdot e(\hat{c}_j, c_j)
\end{equation*}
where $\mathbbm{1}$ is an indicator function, defined as follows:
%
%
\begin{equation*}
    \mathbbm{1}(\hat{c}_j,c) = \begin{cases}
        0 \text{ if }x^\star_j(c) = 1 \text{ and } \hat{c}_j > c_j\\
        0 \text{ if }x^\star_j(c) = 0 \text{ and } \hat{c}_j < c_j\\
        1 \text{ otherwise}
    \end{cases}
\end{equation*}
With this loss component we preserve Property~\ref{prop:zero}, assuming for the base error $e$ the following holds: $e(\hat{c},c) = 0 \iff \hat{c} = c$. For the proof we refer to the Appendix.

The benefit of this component is that the set of predictions s.t.\@ loss is zero is a subset of the set of predictions such that the base loss is $0$, i.e., $\{\hat{c} \in \mathbb{R}^d:\ell^{(\text{O})}(\hat{c},c)=0\} \supset \{\hat{c} \in \mathbb{R}^d:\ell(\hat{c},c)=0\}$ , which means we are penalizing predictions that already cause zero regret less often (as illustrated in Fig.~\ref{fig:asym}).

\subsubsection{Sensitivity-Based One-Sided} 
Based on above principles, we can define another one-sided loss component with weaker properties, but for some problems with higher impact. Linear optimization problems have certain informative properties compared to other optimization problems, among which is sensitivity analysis. 
Given an optimal decision, one can derive for each decision variable the amount of change in the parameter it would require for the optimal decision to change. Let us say our problem is a linear programming problem (LP): $\min_{x \in X} c^T x$ with $X = \{x \in \mathbb{R}^d: Ax \leq b, x \geq 0\}$. Given $c_j$ as the true parameter, we can obtain a lower bound $c^l_j$ and upper bound $c^u_j$ for which $x^\star_j(c)$ remains the same. These are obtained through the optimal basis, which is obtained when solving the LP using the simplex algorithm. This optimal basis consists of linear independent columns of constraint matrix $A$, defining the vertex of the optimal decision. As long as the reduced costs, i.e., rate of change of the objective, of variables that are not in the basis stay positive, the optimal decision does not change. Computing the local bounds $c^l_j, c^u_j$ is a simple linear calculation based on the basis and constraint matrix. With these bounds we can increase the range for which we can set the loss for the predicted parameter to 0:
\begin{equation*}
    \mathbbm{1}_\text{S}(\hat{c}_j,c) = \begin{cases}
        0 \text{ if }x^\star_j(c) = 1 \text{ and } \hat{c}_j > c^l_j\\
        0 \text{ if }x^\star_j(c) = 0 \text{ and } \hat{c}_j < c^u_j\\
        1 \text{ otherwise}
    \end{cases}
\end{equation*}
\begin{equation*}
\ell^{(\text{O}_\text{S})}(\hat{c}, c) = \frac{1}{d}\sum_{j=1}^{d} \mathbbm{1}_\text{S}(\hat{c}_j,c) \hspace{1mm} e(\hat{c}_j, c_j).
\end{equation*}

The guarantee of these bounds not changing the optimal decision only holds locally, i.e., 
only
for a single $\hat{c}_j \in [c_j^l, c_j^u]$, while $\hat{c}_k = c_k$ for $k \neq j$.  Hence when training with this loss, the bounds are almost never a guarantee.  Additionally, this theory only holds for LPs.  To use it for the binary linear problems we tackle in the experiments, we relax the problem to obtain these bounds.  Having no guarantees on the bounds means that Property~\ref{prop:zero} no longer holds: we can have a loss of $0$, without regret actually being $0$.
Nonetheless, for some problems this can be an effective approach. We observe this in the experiments specifically for the travelling salesperson problem, where a particularly cheap edge (two close nodes) can remain optimal even when overpredicting significantly. 


\subsection{Scale-Invariant}

\begin{figure}[tbp]
    \setlength{\abovecaptionskip}{4pt}
    \setlength{\belowcaptionskip}{-4pt}
    \centering
    \includegraphics[width=\linewidth]{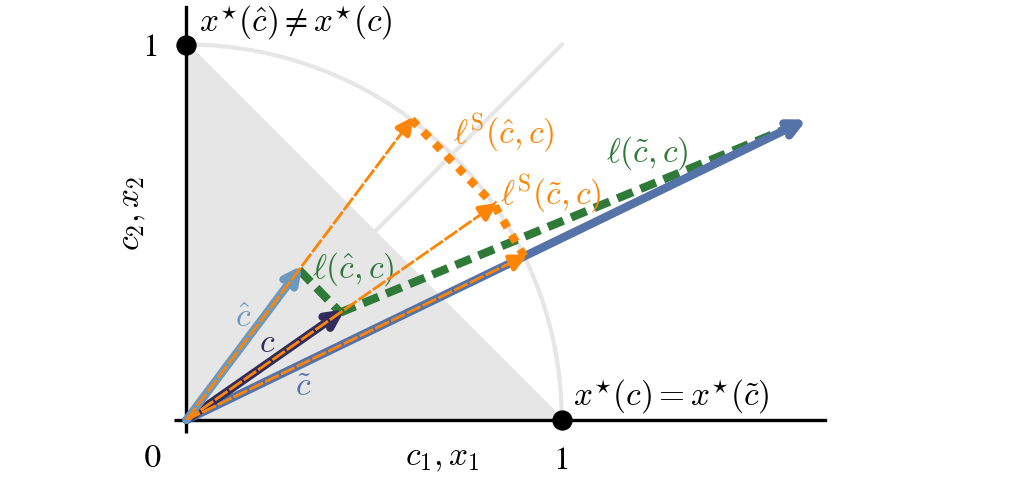}
    \caption{Example of scale-invariant loss on a linear problem $\max_{x \in X} c^T x$ with feasible space (shaded) $X= \{(x_1,x_2): x_1+x_2 \leq 1, x_1 \geq 0, x_2 \geq 0\}$. Shaded line denotes $c_1=c_2$. $\ell$ = $\ell_\text{MSE}$. We observe that $\ell_\text{MSE}(\hat{c},c) \ll \ell_\text{MSE}(\tilde{c},c)$ (green dashed). However, $\hat{c}$ leads to the incorrect decision, while $\tilde{c}$ leads to the correct decision. We also observe that $\ell^{\text{S}}_\text{MSE}(\hat{c},c) \gg \ell^{\text{S}}_\text{MSE}(\tilde{c},c)$ (orange dotted), which better reflects decision quality.} 
    \label{fig:norm}
\end{figure}

Because problem~\eqref{eq:sto} has a linear objective, the optimal decision $x^\star(c)$, and thus the regret, is scale-invariant, i.e., $x^\star(\alpha c) = x^\star(c)$:
\begin{equation*}
    \min_{x \in X} \alpha c^Tx = \alpha \min_{x \in X} c^Tx \implies \\
    \argmin_{x \in X} \alpha c^Tx = \argmin_{x \in X} c^Tx
\end{equation*}
for all $\alpha \in \mathbb{R}^+$. From a cost-sensitive perspective, we want our cost-sensitive loss to share this invariance. We achieve this by applying normalization to both the ground-truth and predicted parameter vectors before evaluating the loss. This normalization is defined as $c^\text{nm} = \frac{c}{\lVert c \rVert}$, where $\lVert c\rVert = \sqrt{\sum_{j=1}^d c_j^2}$.   Since normalization has the property that $c^\text{nm} = (\alpha c)^\text{nm}$ for all $\alpha \in \mathbb{R}^+$, this achieves the desired scale-invariance. Given any loss $\ell$, we can make it scale-invariant (denoted as component `S') by applying per-vector normalization:
\begin{equation*}
    \ell^{(\text{S})}(\hat{c},c) = \ell(\hat{c}^\text{nm}, c^\text{nm}),
\end{equation*}
Adding this component preserves Property~\eqref{prop:zero}, assuming that for $\ell$ the following holds: $\ell(\hat{c},c) = 0 \iff \hat{c} = c$. 

\begin{proof}
Assume $\hat{c}, c \in \mathbb{R}^d$ s.t. $\ell(\hat{c},c) = 0$ based on above assumption we have $\hat{c} = c$. We get: $\hat{c} = c \implies \hat{c}^\text{nm} = c^\text{nm}$, and $\ell_\text{regret}(c^\text{nm}, c^\text{nm})=0$.
\end{proof}

Fig.~\ref{fig:norm} illustrates how MSE and normalized MSE compare, and how the scale-invariance of normalized MSE aligns better with decision quality. By normalizing the parameter vectors, the loss becomes a function of only the angle between these vectors. Because of this the normalized MSE is proportional to the cosine distance. See the Appendix for the proof.

\subsection{Cost-Sensitive Loss}
The loss components together result in a (instance) Cost-based, One-sided and Scale-invariant loss that is COst-Sensitive w.r.t. regret, therefore we name it the COS loss:
\vspace{-1mm}
\begin{equation*}
    \ell^{\text{COS}}_i(\hat{c},c)= C^{(i)} \frac{1}{d}\sum_{j=1}^{d} \mathbbm{1}(\hat{c}_j^\text{nm},c^\text{nm}) \hspace{1mm} e(\hat{c}_j^\text{nm},c_j^\text{nm})
\end{equation*}
\vspace{-1mm}
where $e: \mathbb{R}^d \to \mathbb{R}^d$ is some error function, $c_j^\text{nm}:= (c^\text{nm})_j$. 
There are two things to take into account when combining these components: 1) Instance-based costs are computed based on a baseline predictor, so it is most effective to have this trained with the other components, using loss $\ell^{(\text{O,S})}$. 2) With the scale-invariant component, normalization should be applied before the bounds for either of the one-sided loss components O or $\text{O}_\text{S}$ are determined, as they should be based on the normalized observed vectors $c^\text{nm}$.

\subsubsection{Efficiency}
The computational cost of the different components are summarized in Table~\ref{tab:solves}, in comparison with other DFL approaches that are used in the experiments. We also denote caching and relaxation here to speed up the learning. Relaxation does this by turning an integer LP into a (continuous) LP and is effective when there is a small integrality gap \citep{mipaal}. Caching stores every solve as a feasible decision to create a cache. This is used to find an approximate optimal decision: Since we are dealing with problems with a linear objective, we can compute the dot product $\hat{c}^Tx$ for every $x \in X_\text{cache}$ and take the maximum, which is cheap as long as $X_\text{cache}$ is not too large. CaVE defines cones based on binding constraints in the optimal decisions $x^\star(c)$, requiring determining these upfront and projecting $\hat{c}$ on the optimal cone by solving a quadratic program during training. LAVA is designed for LPs and uses optimality cones in a different fashion: determining all adjacent vertices of optimal decisions, and only penalizing if the prediction favours an adjacent decision over the optimal one. When the problem is not an LP, the adjacent vertices of the linear relaxation are used.

\begin{table}[tbp]
    \setlength{\abovecaptionskip}{4pt}
    \setlength{\belowcaptionskip}{-4pt}
    \centering
    \setlength{\tabcolsep}{2pt}
    \hspace*{-1mm}
    {\smaller\begin{tabular}{ll}
    \toprule
        \textbf{Learning approach \& components} & \textbf{Solver calls} \\
        \midrule

        SPO+, Smart `Predict, then Optimize' \citep{spo}&   $n^\star\!\! + t \! \times \! n$ \\
        PFYL, Perturbed Fenchel Young Loss \citep{perturbed_optimizers}&   $n^\star\!\! + t \! \times \! n$ \\
        MAP, Maximum A Priori \citep{contrastive} &   $n^\star\!\! + t \! \times \! n$ \\
        \hspace{5mm} caching (cache) \citep{contrastive} & $\times p$ \\
        \hspace{5mm} linear relaxation (relax) \citep{mandi2020smart} & $\times r$ \\
        \midrule
                        CaVE, Cone-aligned Vector Estimation \citep{tang2024cave} & $n^\star \!\! +  t\! \times \!n\! \times \!  q$ \\
        LAVA, Loss via Adjacent Vertex Alignment \citep{berden2025} & $n^\star \!\! \times \! r$ \\
        \midrule
        MSE, MAE &  $0$ \\
        \hspace{5mm} instance-based costs (C) &    $+n$ \\
         \hspace{5mm} one-sided (O) &  $+n^\star$ \\
         \hspace{5mm} sensitivity-based one-sided ($\text{O}_\text{S}$) & $+n^\star \!\!\times r$  \\
         \hspace{5mm} scale-invariant (S) &  $+0$ \\
  
    \bottomrule
    \end{tabular}}
    \caption{Number of solver calls for different DFL approaches, incl.\@ components to improve efficiency/performance. Compared with accuracy-based losses with our presented cost-sensitive components.  $n$: size of dataset, $t$: number of epochs run, $n^\star$: optimal decisions on the dataset, which is $0$ when these are known upfront and $n$ otherwise. Ratios $p$, $q$, $r$ are expected to lie in $[0,1]$; $p$: percentage of solves done when using caching, $r$ \& $q$: discount for solving the relaxed model, or a quadratic projection.}
    \label{tab:solves}
\end{table}

\section{Experimental Evaluation}
\label{sec:results}

We perform two sets of experiments to address 
these questions:
\begin{itemize}[leftmargin=*,nosep]
    \item Do the presented instance-based costs outperform the proposed weighting by \citet{lawless2022}?
    \item What is the impact of the different COS loss components? 
    \item How effective is the sensitivity-based one-sided loss? \item How well does mean absolute error (MAE) work compared to MSE as a base loss?
    \item How far can we scale compared to other presented scalable DFL approaches?
\end{itemize}
We refer to the Appendix for details and exhaustive results. Code was implemented using \textit{PyDFLT} \citep{PyDFLT2026}.

\subsection{Component-Wise Analysis}
The first set of experiments aims to evaluate the loss-components.  We explore three commonly-used binary LPs: the multi-dimensional knapsack with 32 items (KS32), shortest path on a 5 by 5 grid (SP5x5) and travelling salesperson problem with 20 nodes (TSP20), as introduced by \citet{spo} and studied more extensively by \citet{pyepo}. For these problems correlated features and uncertain parameters are generated to create a predict-then-optimize problem. SPO+ is used as SOTA comparison. 

\begin{figure}[tbp]
    \setlength{\abovecaptionskip}{4pt}
    \setlength{\belowcaptionskip}{-4pt}
    \centering
    \includegraphics[width=\linewidth,clip,trim={0 0 0 0mm}]{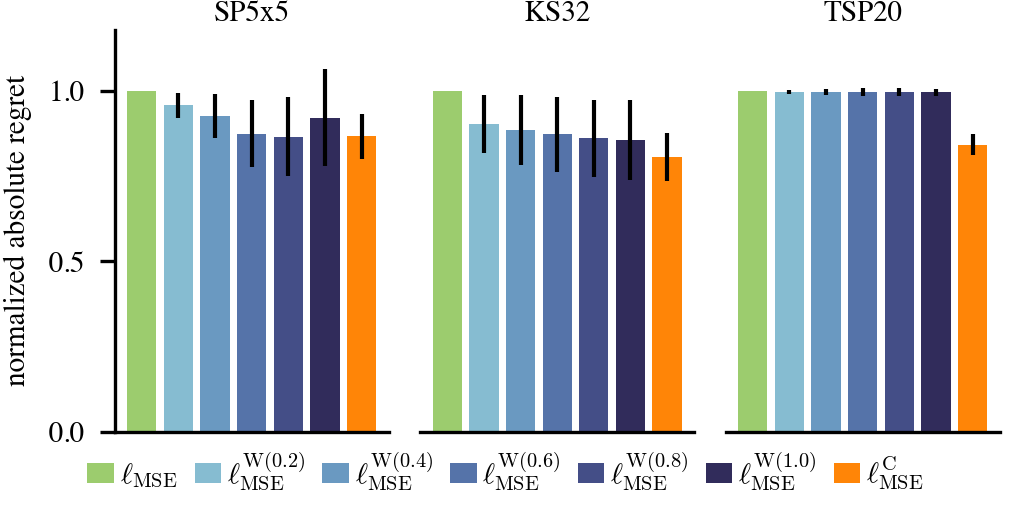}
    \caption{Comparing our instance-based costs loss $\ell_\text{MSE}^{\text{C}}$ with $\ell_\text{MSE}^{\text{W}(w)}$ for $w \in \{0, 0.2, 0.4, 0.6, 0.8, 1.0\}$. Test absolute regret normalized on $\ell_\text{MSE}(=\ell_\text{MSE}^{\text{W}(0.0)})$. Based on 20 seeds. Error bars denote stdev.}
    \label{fig:instance-based}
\end{figure}

\paragraph{Instance-based costs}
To validate the instance-based costs based on a local linear relationship with regret, we compare with the approach by \citet{lawless2022}, Equation~\eqref{eq:lawless}, with different values for hyperparameter $w$. In Fig.~\ref{fig:instance-based} we observe that our proposed instance-based costs consistently perform at least as good as the loss from \citeauthor{lawless2022}.

\begin{figure}[tbp]
    \setlength{\abovecaptionskip}{2pt}
    \setlength{\belowcaptionskip}{-4pt}
    \centering
    \includegraphics[width=\linewidth,clip,trim={0 0 0 0mm}]{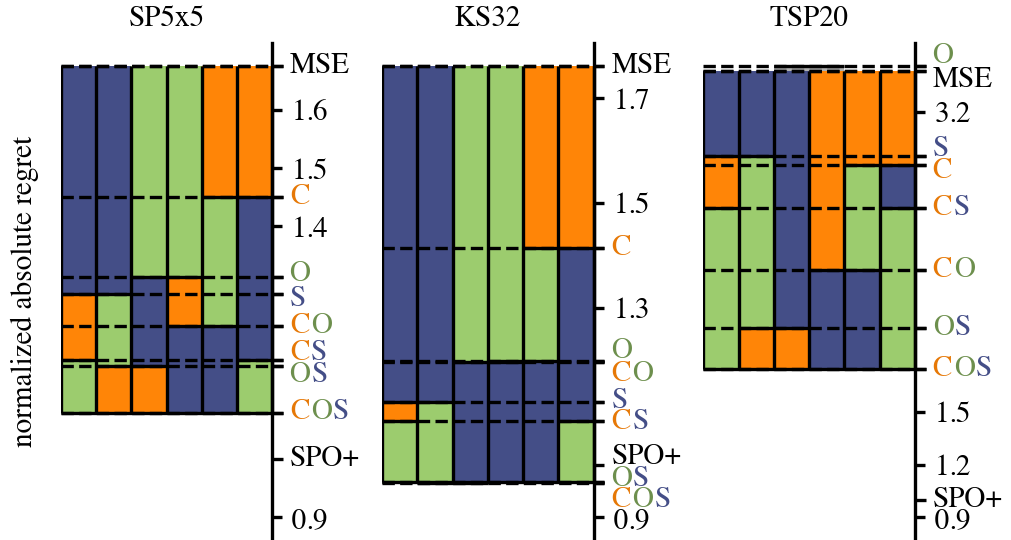}
    \caption{Test absolute regret normalized on SPO+ for all loss component combinations with MSE. Each column represents an order of adding the components (e.g., left-most is S $\to$ C $\to$ O). Colours denote the improvement based on adding a component: Scale-invariant (S, blue), one-sided (O, green) and instance-based costs (C, orange). Base loss MSE and SPO+ added for comparison. Total area per colour indicates total added value.}
    \label{fig:monotonicity}
\end{figure}

\paragraph{Impact of COS loss components}
We analyze how much test regret is improved by each of three main loss components: instance-based costs (C), one-sided loss (O) and scale-invariance (S). To do this, we run every combination on the benchmarks with $e$ being the squared error, resulting in MSE without any components. By plotting the incremental improvements from each added component in Fig.~\ref{fig:monotonicity}, we observe that: 1) Adding components is empirically monotonic, i.e., for every additional component performance improves or stays the same. 2) Based on the surface area of each colour (component), scale-invariance (S) has most impact followed by the one-sided loss (O). Since the instance-based costs component is also the most expensive (see Table~\ref{tab:solves}), using just the other two components is a valid choice when scaling up.

\begin{figure}[tbp]
    \setlength{\abovecaptionskip}{4pt}
    \setlength{\belowcaptionskip}{-4pt}
    \centering
    \includegraphics[width=\linewidth,clip,trim={0 0 0 0mm}]{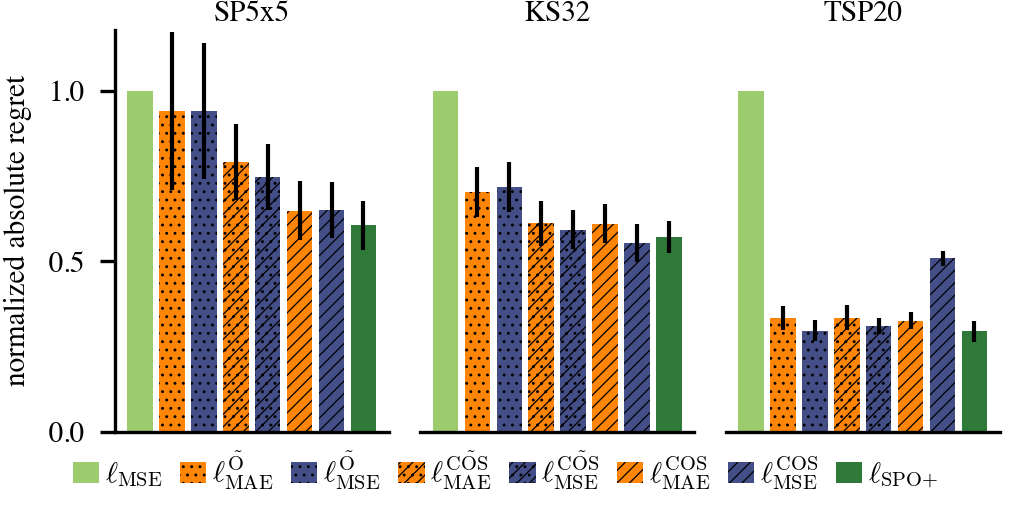}
    \caption{Test absolute regret normalized on $\ell_\text{MSE}$: Comparing COS, C$\tilde{\text{O}}$S and $\tilde{\text{O}}$, with $\ell_\text{MAE}$ and $\ell_\text{MSE}$.}
    \label{fig:asmae}
\end{figure}

\paragraph{Sensitivity-based one-sided loss $\text{O}_\text{S}$ and MAE}
Sect.~\ref{sec:asym} presented an asymmetric one-sided loss based on sensitivity analysis $(\text{O}_\text{S})$.  Due to the sensitivity bounds being based on local single parameter adjustments, the component violates Property~\eqref{prop:zero}, and its effectiveness is unclear.
Also, we have presented MSE as a valid choice for a base loss as it has the Property~\eqref{prop:zero} and is a commonly used accuracy metric. Here we evaluate the performance of mean absolute error (MAE) $\ell_\text{MAE}(\hat{c},c) = \frac{1}{d}\sum_{j=1}^d|\hat{c}-c|$, which is another common loss function that penalizes deviations from the true value linearly, instead of quadratically with MSE, penalizing outliers more. However, it is unclear which metric has a stronger correlation to regret in general. Fig.~\ref{fig:asmae} shows the most relevant component combinations: (C,O,S), (C,$\text{O}_\text{S}$,S) and $(\text{O}_\text{S})$ on both MSE and MAE. We observe: 1) Similar performance to SOTA $\ell_\text{SPO+}$ for multiple combinations. Specifically $\ell_\text{MAE}^{(\text{C,O,S})}$ performs consistently similar to $\ell_\text{SPO+}$. 2) Component $\text{O}_\text{S}$ is especially effective on the TSP, already performing better than other component combinations by itself. 

\begin{figure*}[t]
    \setlength{\abovecaptionskip}{4pt}
    \setlength{\belowcaptionskip}{-2pt}
    \centering
    \includegraphics[width=\linewidth,clip,trim={0mm 0mm 0 0}]{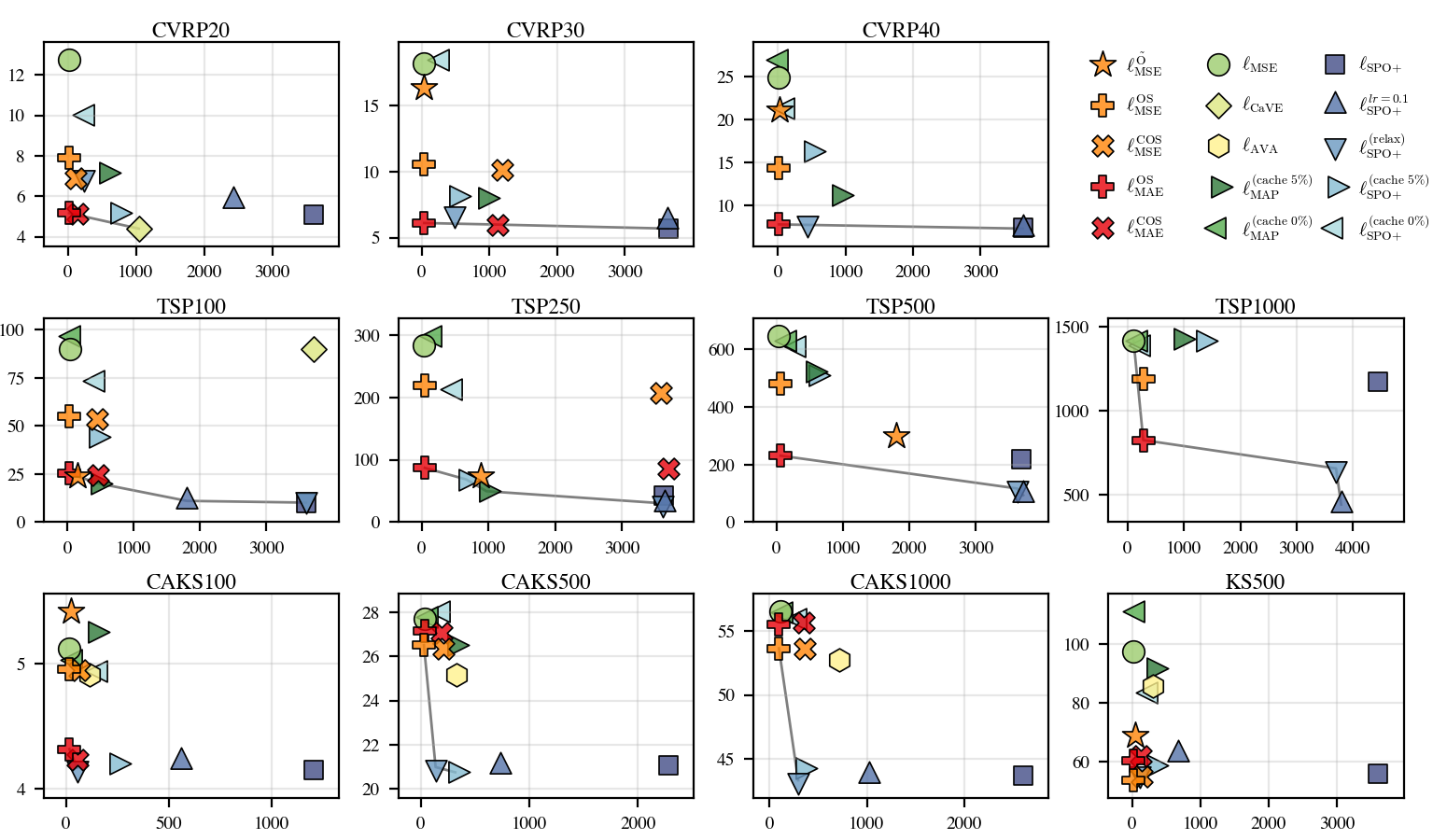}
    \caption{Test absolute regret (y-axis) and runtime in seconds (x-axis) based on 3 seeds. Losses with regret $\geq 15\%$ worse than $\ell_\text{MSE}$ are excluded. Grey lines denote the Pareto front: losses such that no other loss has lower-or-equal regret and runtime (within 30s considered equal).}
    \label{fig:results}
\end{figure*}

\subsection{Scalable Decision-Focused Learning}
The second of the experiments aims to evaluate performance in decision quality as well as runtime, highlighting the benefit of our proposed approach.  We explore three problems: TSP but now with size 100, 250, 500 and 1000; capacitated vehicle routing problem (CVRP) as presented in \citet{tang2024cave}, with 20, 30 and 40 nodes; California house price data-based multi-dimensional knapsack (CAKS), where the goal is to buy the most valuable houses given some multi-dimensional budget \citep{berden2025}. We compare with an extensive set of approaches, denoted in Table~\ref{tab:solves}. We assume optimal decisions on the training set $x^\star(c^{(i)})$ for all $i$ are known.


Based on our first set of experiments we choose five component combinations. 
Figure~\ref{fig:results} denotes all methods we compare with, including components like relaxation or caching. For caching we include $p=0\%$, requiring no solves. All methods use a learning rate of 0.01, except $\ell_\text{SPO+}^{\text{lr}=0.1}$, allowing faster learning. Convergence is determined by a validation set: when there has been 5 minutes without improvement, training is terminated. Each approach runs at most 100 epochs. Except for SPO+, 90\% of the train/validation set was used for training and 10\% for validation. Validation is always done without any solves, i.e., the validation is done based on the loss-function evaluation on the validation set, as using solves in validation would significantly slow down the training. For SPO+, the full train/validation set is used for training, with validation being the training loss. In the case caching is used, validation is done on the cache. A time-out of one hour is used.

In Fig.~\ref{fig:results} we observe strong performance 
of our losses, both in decision quality and computation time. Some SPO+ variants achieve better decision quality, but mostly with longer run-time. In general COS loss with MAE shows high quality, while having minimal runtime. Some approaches work well for a certain problem: relaxed SPO+ is really effective for all KS problems as they have a small integrality gap. 

Our methods seem to perform poorly on CAKS500 and CAKS1000. We believe this might stem from the fact that in this problem there is a single shared model that predicts for each uncertain parameter separately, i.e., there is a set of features associated with each parameter, instead of having global features that map directly to all parameters.  Because of this we ran KS500, a knapsack problem with the same generation of weights, but generated features that map as a whole to the uncertain parameters. Here we see that our approach performs well, indicating that performance relates to the local features in combination with the problem size.

\section{Discussion and Related Work}
The cost-sensitive multi-output regression framework shows to be highly effective for DFL. Multi-output regression has been studied before in a cost-sensitive setting by \citet{peng2023}, applying it to imbalanced data streams to obtain stable predictions. DFL specifically is considered in a cost-sensitive setting: \citet{vanderschueren2022} compared predict-then-optimize to DFL in classification. Regarding the instance-based costs, \citet{decorte2023} present cost-sensitive regression through one-step boosting, showing resemblance with our approach. However, the boosting step requires the true loss function to be differentiable, which is not the case in our setting. Another related work casts cost-sensitive classification as a one-sided (single-output) regression problem \citep{tu2010}.

To improve DFL efficiency, there is a line of work that uses neural networks to replace the solver, mostly requiring a procedure to ensure feasibility. This includes \citep{kotary2024learning}, which is aimed at continuous optimization, and which uses a neural solver to produce solutions directly from features, without the intermediate cost prediction layer. For arbitrary problems, \citet{schutte2025} present different effective solver approximations. Also, \citet{mckenzie2023differentiating} use Jacobian-free backpropagation to efficiently backpropagate through the neural solver, which specifically works for (I)LPs. Finally, several works try learning surrogate losses upfront, to then be used in training \citep{lodl,leaving_the_nest}. Despite being efficient at training time, a high number of solves is required to find an effective surrogate loss. Our approach requires zero or one solve per training instance.

\section{Conclusions}
\label{sec:conc}

The contribution of this work has been to cast decision-focused learning in the lens of cost-sensitive learning, namely therein as a multi-output regression problem. As a result we could propose new loss components which lead to marked improvements in DFL efficiency, allowing scaling to more complex problems. Future work includes further analyzing problem settings with local features, as well as applying the proposed framework on more problems from practice.


\FloatBarrier
\bibliographystyle{named}
\bibliography{our-refs}

@inproceedings{contrastive,
  title={Contrastive Losses and Solution Caching for Predict-and-Optimize},
  author={Mulamba, Maxime and Mandi, Jayanta and Diligenti, Michelangelo and Lombardi, Michele and Bucarey, Victor and Guns, Tias and others},
  booktitle={Proceedings of the Thirtieth International Joint Conference on Artificial Intelligence},
  pages={2833--2840},
  year={2021},
  noorganization={ijcai. org}
}

@article{dfl_survey,
  title={Decision-focused learning: Foundations, state of the art, benchmark and future opportunities},
  author={Mandi, Jayanta and Kotary, James and Berden, Senne and Mulamba, Maxime and Bucarey, Victor and Guns, Tias and Fioretto, Ferdinando},
  journal={Journal of Artificial Intelligence Research},
  volume={80},
  pages={1623--1701},
  year={2024}
}

@article{melding,
title={Melding the Data-Decisions Pipeline: Decision-Focused Learning for Combinatorial Optimization}, volume={33}, url={https://ojs.aaai.org/index.php/AAAI/article/view/3982}, DOI={10.1609/aaai.v33i01.33011658}, number={01}, journal={Proceedings of the AAAI Conference on Artificial Intelligence}, author={Wilder, Bryan and Dilkina, Bistra and Tambe, Milind}, year={2019}, nomonth={Jul.}, pages={1658-1665}
}

@inproceedings{perturbed_optimizers,
 author = {Berthet, Quentin and Blondel, Mathieu and Teboul, Olivier and Cuturi, Marco and Vert, Jean-Philippe and Bach, Francis},
 booktitle = {Advances in Neural Information Processing Systems},
 editor = {H. Larochelle and M. Ranzato and R. Hadsell and M.F. Balcan and H. Lin},
 pages = {9508--9519},
 publisher = {Curran Associates, Inc.},
 title = {Learning with Differentiable Pertubed Optimizers},
 url = {https://proceedings.neurips.cc/paper_files/paper/2020/file/6bb56208f672af0dd65451f869fedfd9-Paper.pdf},
 novolume = {33},
 year = {2020}
}

@article{spo,
author = {Elmachtoub, Adam N. and Grigas, Paul},
title = {Smart `Predict, then Optimize'},
journal = {Management Science},
volume = {68},
number = {1},
pages = {9-26},
year = {2022},
doi = {10.1287/mnsc.2020.3922},
URL = { 
        https://doi.org/10.1287/mnsc.2020.3922
},
eprint = { 
        https://doi.org/10.1287/mnsc.2020.3922
}
}

@inproceedings{pyepo,
  title={PyEPO: A PyTorch-based End-to-End Predict-then-Optimize Library with Linear Objective Function},
  author={Tang, Bo and Khalil, Elias B},
  booktitle={OPT 2022: Optimization for Machine Learning (NeurIPS 2022 Workshop)},
  year={2022}
}

@misc{PyDFLT2026,
    title = {{PyDFLT: A} {Python}-based Decision-Focused Learning Toolbox},
    author = {Noah Schutte and Kim van den Houten and Grigorii Veviurko},
    howpublished = {\url{https://doi.org/10.5281/zenodo.20177131}},
    year = {2026}
}

@inproceedings{mipaal,
  title={{MIPaaL}: Mixed integer program as a layer},
  author={Ferber, Aaron and Wilder, Bryan and Dilkina, Bistra and Tambe, Milind},
  booktitle={Proceedings of the AAAI Conference on Artificial Intelligence},
  novolume={34},
  nonumber={02},
  pages={1504--1511},
  year={2020}
}

@inproceedings{lodl,
  author       = {Sanket Shah and
                  Kai Wang and
                  Bryan Wilder and
                  Andrew Perrault and
                  Milind Tambe},
  noeditor       = {Sanmi Koyejo and
                  S. Mohamed and
                  A. Agarwal and
                  Danielle Belgrave and
                  K. Cho and
                  A. Oh},
  title        = {Decision-Focused Learning without Decision-Making: Learning Locally
                  Optimized Decision Losses},
  fullbooktitle    = {Advances in Neural Information Processing Systems 35: Annual Conference
                  on Neural Information Processing Systems 2022, NeurIPS 2022, New Orleans,
                  LA, USA, November 28 - December 9, 2022},
  booktitle = {Proceedings of NeuIPS},
  year         = {2022},
  url          = {http://papers.nips.cc/paper\_files/paper/2022/hash/0904c7edde20d7134a77fc7f9cd86ea2-Abstract-Conference.html},
  bibsource    = {dblp computer science bibliography, https://dblp.org}
}

@inproceedings{leaving_the_nest,
  title={Leaving the nest: Going beyond local loss functions for predict-then-optimize},
  author={Shah, Sanket and Wilder, Bryan and Perrault, Andrew and Tambe, Milind},
  booktitle={Proceedings of the AAAI Conference on Artificial Intelligence},
  novolume={38},
  nonumber={13},
  pages={14902--14909},
  year={2024}
}

@inproceedings{mandi2020smart,
  title={Smart predict-and-optimize for hard combinatorial optimization problems},
  author={Mandi, Jayanta and Stuckey, Peter J and Guns, Tias and others},
  booktitle={Proceedings of the AAAI Conference on Artificial Intelligence},
  novolume={34},
  nonumber={02},
  pages={1603--1610},
  year={2020}
}

@inproceedings{tang2024cave,
  author       = {Bo Tang and
                  Elias B. Khalil},
  editor       = {Bistra Dilkina},
  title        = {CaVE: {A} Cone-Aligned Approach for Fast Predict-then-optimize with
                  Binary Linear Programs},
  fullbooktitle    = {Integration of Constraint Programming, Artificial Intelligence, and
                  Operations Research - 21st International Conference, {CPAIOR} 2024,
                  Uppsala, Sweden, May 28-31, 2024, Proceedings, Part {II}},
  booktitle = {Procedings of CPAIOR},
  series       = {Lecture Notes in Computer Science},
  volume       = {14743},
  pages        = {193--210},
  publisher    = {Springer},
  year         = {2024},
  url          = {https://doi.org/10.1007/978-3-031-60599-4\_12},
  doi          = {10.1007/978-3-031-60599-4\_12},
  bibsource    = {dblp computer science bibliography, https://dblp.org}
}

@inproceedings{kotary2024learning,
  author       = {James Kotary and
                  Vincenzo Di Vito and
                  Jacob Christopher and
                  Pascal Van Hentenryck and
                  Ferdinando Fioretto},
  noeditor       = {Ulle Endriss and
                  Francisco S. Melo and
                  Kerstin Bach and
                  Alberto Jos{\'{e}} Bugar{\'{\i}}n Diz and
                  Jose Maria Alonso{-}Moral and
                  Sen{\'{e}}n Barro and
                  Fredrik Heintz},
  title        = {Learning Joint Models of Prediction and Optimization},
  fullbooktitle    = {{ECAI} 2024 - 27th European Conference on Artificial Intelligence,
                  19-24 October 2024, Santiago de Compostela, Spain - Including 13th
                  Conference on Prestigious Applications of Intelligent Systems {(PAIS}
                  2024)},
  booktitle = {Proceedings of ECAI},
  noseries       = {Frontiers in Artificial Intelligence and Applications},
  novolume       = {392},
  pages        = {2476--2483},
  nopublisher    = {{IOS} Press},
  year         = {2024},
  url          = {https://doi.org/10.3233/FAIA240775},
  doi          = {10.3233/FAIA240775},
  bibsource    = {dblp computer science bibliography, https://dblp.org}
}

@article{mckenzie2023differentiating,
  author       = {Daniel McKenzie and
                  Howard Heaton and
                  Samy Wu Fung},
  title        = {Differentiating Through Integer Linear Programs with Quadratic Regularization
                  and {D}avis-{Y}in Splitting},
  journal      = {Trans. Mach. Learn. Res.},
  volume       = {2024},
  year         = {2024},
  url          = {https://openreview.net/forum?id=H8IaxrANWl},
  bibsource    = {dblp computer science bibliography, https://dblp.org}
}

@article{decorte2023,
title = {Interpretable cost-sensitive regression through one-step boosting},
journal = {Decision Support Systems},
volume = {175},
pages = {114024},
year = {2023},
issn = {0167-9236},
doi = {https://doi.org/10.1016/j.dss.2023.114024},
url = {https://www.sciencedirect.com/science/article/pii/S0167923623000994},
author = {Thomas Decorte and Jakob Raymaekers and Tim Verdonck}
}

@article{vanderschueren2022,
title = {Predict-then-optimize or predict-and-optimize? {A}n empirical evaluation of cost-sensitive learning strategies},
journal = {Information Sciences},
volume = {594},
pages = {400-415},
year = {2022},
issn = {0020-0255},
doi = {https://doi.org/10.1016/j.ins.2022.02.021},
url = {https://www.sciencedirect.com/science/article/pii/S0020025522001542},
author = {Toon Vanderschueren and Tim Verdonck and Bart Baesens and Wouter Verbeke}
}

@inproceedings{elkan2001,
  title={The foundations of cost-sensitive learning},
  author={Elkan, Charles},
  booktitle={International Joint Conference on Artificial Intelligence},
  novolume={17},
  nonumber={1},
  pages={973--978},
  year={2001},
  noorganization={Lawrence Erlbaum Associates Ltd}
}

@article{granger1969,
  title={Prediction with a generalized cost of error function},
  author={Granger, Clive WJ},
  journal={Journal of the Operational Research Society},
  volume={20},
  number={2},
  pages={199--207},
  year={1969},
  publisher={Taylor \& Francis}
}

@article{lawless2022,
  title={A note on task-aware loss via reweighing prediction loss by decision-regret},
  author={Lawless, Connor and Zhou, Angela},
  journal={arXiv preprint arXiv:2211.05116},
  year={2022}
}

@article{sadana2025,
title = {A survey of contextual optimization methods for decision-making under uncertainty},
journal = {European Journal of Operational Research},
volume = {320},
number = {2},
pages = {271-289},
year = {2025},
issn = {0377-2217},
doi = {https://doi.org/10.1016/j.ejor.2024.03.020},
url = {https://www.sciencedirect.com/science/article/pii/S0377221724002200},
author = {Utsav Sadana and Abhilash Chenreddy and Erick Delage and Alexandre Forel and Emma Frejinger and Thibaut Vidal}
}

@article{berden2025,
  title={Solver-Free Decision-Focused Learning for Linear Optimization Problems},
  author={Berden, Senne and Mahmuto{\u{g}}ullar{\i}, Ali {\.I}rfan and Tsouros, Dimos and Guns, Tias},
  journal={arXiv preprint arXiv:2505.22224},
  year={2025}
}

@article{zhao2011,
  title={An extended tuning method for cost-sensitive regression and forecasting},
  author={Zhao, Huimin and Sinha, Atish P and Bansal, Gaurav},
  journal={Decision Support Systems},
  volume={51},
  number={3},
  pages={372--383},
  year={2011},
  publisher={Elsevier}
}

@article{zellner1986,
  title={Bayesian estimation and prediction using asymmetric loss functions},
  author={Zellner, Arnold},
  journal={Journal of the American Statistical Association},
  volume={81},
  number={394},
  pages={446--451},
  year={1986},
  publisher={Taylor \& Francis}
}

@article{bansal2008,
  title={Tuning data mining methods for cost-sensitive regression: A study in loan charge-off forecasting},
  author={Bansal, Gaurav and Sinha, Atish P and Zhao, Huimin},
  journal={Journal of Management Information Systems},
  volume={25},
  number={3},
  pages={315--336},
  year={2008},
  publisher={Taylor \& Francis}
}

@article{peng2023,
  title={Multi-output regression for imbalanced data stream},
  author={Peng, Tao and Sellami, Sana and Boucelma, Omar and Chbeir, Richard},
  journal={Expert Systems},
  volume={40},
  number={10},
  pages={e13417},
  year={2023},
  publisher={Wiley Online Library}
}

@inproceedings{tu2010,
  title={One-sided Support Vector Regression for Multiclass Cost-sensitive Classification.},
  author={Tu, Han-Hsing and Lin, Hsuan-Tien},
  booktitle={ICML},
  number={4},
  pages={5},
  year={2010}
}

@article{schutte2025,
  title={Sufficient Decision Proxies for Decision-Focused Learning},
  author={Schutte, Noah and Veviurko, Grigorii and Postek, Krzysztof and Yorke-Smith, Neil},
  journal={arXiv preprint arXiv:2505.03953},
  year={2025}
}

\FloatBarrier
\newpage
\clearpage

\appendix

\section{One-sided loss general case}
In the main work we discuss the asymmetric loss for the case where the problem is a binary LP. In a general case, assuming only a linear objective and bounded feasible space, we can apply the same logic.  Boundedness implies that there exist $l, u$ such that $l \leq x \leq u$ for all $x \in X$.  Given these bounds, the penalization function generalizes to the elements of the optimal decisions attaining these bounds.  However, these bounds are not guaranteed to be attained (except in a binary problem)
and are likely loose.

\begin{equation}
    \mathbbm{1}(\hat{c}_j,c_j) = \begin{cases}
        0 \text{ if }x^\star_j(c) = u_j \text{ and } \hat{c}_j > c_j\\
        0 \text{ if }x^\star_j(c) = l_j \text{ and } \hat{c}_j < c_j\\
        1 \text{ otherwise}
    \end{cases}
\end{equation}

\section{Proofs}
\subsection{Regret consistency of one-sided loss}
\begin{proof}
Assume a binary LP, with feasibility space $X = \{0,1\}^d$, and a base error $e$ such that
\[
e(\hat c_j,c_j)=0 \iff \hat c_j=c_j.
\]
By definition of $\ell^{(A)}$:
\[
\ell^{(A)}(\hat c,c)=0 
\iff 
e(\hat c_j,c_j)=0 \;\lor\; \mathbbm{1}(\hat c_j,c)=0 
\quad \forall j.
\]
So, if $\ell^{(A)}(\hat c,c)=0$, this implies that for all $j$:
$$\hat c_j - c_j \geq 0 \text{ if }x_j^\star(c)=1 \text{ and } \hat c_j - c_j \leq 0 \text{ if } x_j^\star(c)=0$$
Now for any feasible $x\in X$. We have for each $j$: \\
If $x_j^\star(c)=1$, then $x_j-x_j^\star(c)\le 0$ and $\hat c_j-c_j\ge 0$. \\
If $x_j^\star(c)=0$, then $x_j-x_j^\star(c)\ge 0$ and $\hat c_j-c_j\le 0$. \\
This means that 
\[
(\hat c-c)^\top(x-x^\star(c))\le 0
\quad \forall x\in  X.
\]
Resulting in 
\begin{equation*}
\hat c^\top x - \hat c^\top x^\star(c)
\;\le\;
c^\top x - c^\top x^\star(c) \leq 0
\quad \forall x \in X,
\end{equation*}
where the last inequality follows from $x^\star(c)$ being optimal. So arranging the left hand side we get:
\[
\hat c^\top x \le \hat c^\top x^\star(c)
\quad \forall x\in  X,
\]
which shows that $x^\star(c)$ obtains a better objective w.r.t. any $\hat{c}$ and is therefore optimal. Thus,
\[
x^\star(\hat c)=x^\star(c)
\quad 
\]
Finally, consider the regret:
\[
\ell_{\mathrm{regret}}(\hat c,c)
= c^\top x^\star(c)-c^\top x^\star(\hat c) = 0.
\]
So $\ell^{(A)}$ is regret consistent.
\end{proof}

\subsection{Normalized MSE proportional to cosine distance}
\begin{proof}
Let $c, \hat{c} \in \mathbb{R}^d$ be normalized vectors such that $\|c\| = \|\hat{c}\| = 1$, and cosine distance is defined as: $1- c \cdot \hat{c}$. 
\begin{align*}
\ell_\text{MSE}(\hat{c},c)&=\frac{1}{d} \|c - \hat{c}\|^2 = (c - \hat{c}) \cdot (c - \hat{c})\\
&= \frac{1}{d}( c \cdot c - 2(c \cdot \hat{c}) + \hat{c} \cdot \hat{c} )
\end{align*}

\noindent
Normalized vectors:
$$ = \frac{1}{d}(1 - 2(c \cdot \hat{c}) + 1 ) = \frac{2}{d} (1 - c \cdot \hat{c})$$

Thus, the normalized MSE is directly proportional to the cosine distance with a proportionality constant of $\frac{2}{d}$.
\end{proof}

\section{Iterative instance-based costs}

\citet{lawless2022} also explore the idea of an iterative approach: After obtaining our final predictive model, we can re-compute the instance-based costs and train with these new costs. This approach, with our instance-based costs, is shown in Algorithm~\ref{alg:1}. In their experiments they show that this iterative approach does not work significantly better than a single determination of the costs. We think this can be explained as follows: In the first iteration, we weight the instances based on how well or poorly they perform w.r.t.\@ regret, such that the loss gets redistributed relative to this regret. This makes poorly performing instances perform better, and the other way around. If we now reweight based on our new predictive model, it is going to put more weight on the now poor performing instances, working against our first reweighting.

\begin{algorithm}
  \caption{Iterative Instance-Based Costs}
  \begin{algorithmic}[1]
      \State \textbf{Input:} Dataset $\mathcal{D} = \{z^{(i)}, c^{(i)}\}_{i=1}^n$, base loss $\ell$, iterations $K$, $C_1^{(i)} \gets 1$ for all $i=1, \dots, n$
      \For{$k = 1$ \textbf{to} $K$}
          \State $\hat{\theta}_k \gets \argmin_\theta \sum_{i=1}^n C_k^{(i)}\ell(\phi_\theta(z),c)$
          \State $\bar{c}_k \gets \phi_{\hat{\theta}_k}(z)$
          \State $C^{(i)}_{k+1} \gets \frac{\ell_{\text{regret}}(\bar{c}^{(i)}_k, c^{(i)})}{\ell(\bar{c}^{(i)}_k,c^{(i)})}$, for all $i \in \{1, \dots, n\}$
      \EndFor

      \State \textbf{Output:} predictive model $\phi_{\hat{\theta}_K}$
  \end{algorithmic} \label{alg:1}
  \end{algorithm}

\begin{algorithm}
\caption{Ensemble Iterative Instance-Based Costs}
\begin{algorithmic}[1]
    \State \textbf{Input:} Dataset $\mathcal{D} = \{z^{(i)}, c^{(i)}\}_{i=1}^n$, base loss $\ell$, iterations $K$, $C_1^{(i)} \gets 1$ for all $i=1, \dots, n$    
    \For{$k = 1$ \textbf{to} $K$}
        \State $\hat{\theta}_k \gets \argmin_\theta \sum_{i=1}^n C_k^{(i)}\ell(\phi_\theta(z),c)$ 
        \State $\Phi_k(z) \gets \frac{1}{k} \sum_{i=1}^k \phi_{\hat{\theta}_i}(z)$ 
        \State $\bar{c}_k \gets \Phi_k(z)$
        \State $C^{(i)}_{k+1} \gets \frac{\ell_{\text{regret}}(\bar{c}^{(i)}_k, c^{(i)})}{\ell(\bar{c}^{(i)}_k,c^{(i)})}$, for all $i \in \{1, \dots, n\}$ 
    \EndFor
    
    \State \textbf{Output:} Ensemble model $\Phi_K$
\end{algorithmic} \label{alg:2}
\end{algorithm}

\begin{figure}[tbp]
    \centering
    \includegraphics[width=\linewidth,clip,trim={0 0 0 0mm}]{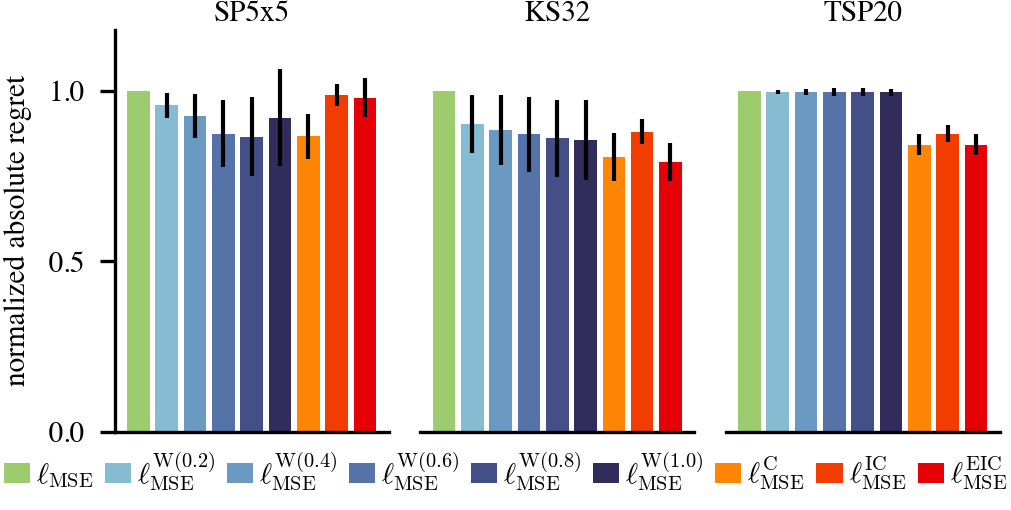}
    \caption{Comparing our instance-based costs loss $\ell_\text{MSE}^{\text{C}}$, iterative instance-based costs $\ell_\text{MSE}^{\text{IC}}$, ensemble iterative instance-based costs $\ell_\text{MSE}^{\text{EIC}}$ and $\ell_\text{MSE}^{\text{W}(w)}$ for $w \in \{0, 0.2, 0.4, 0.6, 0.8, 1.0\}$. Test absolute regret normalized on $\ell_\text{MSE}(=\ell_\text{MSE}^{\text{W}(0.0)})$. Based on 20 seeds. Error bars denote stdev.}
    \label{fig:instance-based-reweigh}
\end{figure}

To counter this, we try an iterative scheme that does not discard each earlier determined predictive model. Inspired by boosting, where an ensemble of predictive models is used, we combine each predictive model obtained at each iteration into an ensemble, recomputing the instance-based costs after every ensemble update. This is presented in Algorithm~\ref{alg:2}. Figure~\ref{fig:instance-based-reweigh} shows experimental results based on these to iterative approaches. Despite being comparable with a single iteration in some cases, we consider these approaches as ineffective. This is because every iteration requires an a solve call for every training instance, reducing efficiency without significant performance gains.

\section{Experimental Details and Results}

\subsubsection{Benchmark Experiments}

The full results for the first set of experiments are shown in Table~\ref{tab:bench}. For the data generation for these problems we set polynomial degree 6, noise width 0.5. The shortest path is 5x5, TSP 20 nodes and knapsack two-dimensional with 32 items and a capacity of 20 (as in PyEPO). For the knapsack the item weights are distributed uniformly from 3 to 8, being the only problem with randomness in the problem formulation. Because of this and the randomness in the data generation all experiments were ran on 20 seeds. Datasets are used with a 1000, 400, 600 train/validation/test split. The validation is used to pick the best observed model over all epochs. Learning rates are set to 0.005, and every approach runs 50 epochs which is plenty for convergence.

\subsubsection{Scalability Experiments}
The full results for the second set of experiments are shown in Table~\ref{tab:scale}. For the CVRP, 5 vehicles are used with a capacity of 30. Demand is generated uniformly from 1 to 10, as in \citet{tang2024cave}. One adjustment is that we scale demand such that the problem is always feasible, by making total demand 70\% of the total capacity of all vehicles together (this is about the average ratio in CVRP20, while the larger sizes have a high likelihood to be infeasible without scaling).
For the housing problem we use the same setup as \citet{berden2025}. There are 8 features. The knapsack is 4 dimensional and we use randomly generated weight between 3 and 8. Capacity at each dimension is taken to be 10\% of the sum of all weights for that dimension. For the TSP problems and KS500, data is generated as in the first set of experiments.  
For dataset sizes and number of seeds see table \ref{tab:scale}. We assume optimal decisions $x^\star(c^{(i)})$ and optimal decision relaxations $x_\text{relax}^\star(c^{(i)})$ are known for all $i$. The initial computation times for CaVE, LAVA are not included.

\begin{table}[t]
\centering
\begingroup\footnotesize
\setlength{\tabcolsep}{5pt}
\renewcommand{\arraystretch}{1.2}
\begin{tabular}{lccc}
\toprule
 & SP5x5 & KS32 & TSP20 \\
\midrule
$\ell_\text{MSE}$ & 1.00 (0.00) & 1.00 (0.00) & 1.00 (0.00) \\
$\ell_\text{MSE}^{\text{(C)}}$ & 0.86 (0.07) & 0.79 (0.07) & 0.84 (0.03) \\
$\ell_\text{MSE}^{\text{(O)}}$ & 0.79 (0.09) & 0.68 (0.06) & 1.02 (0.18) \\
$\ell_\text{MSE}^{\text{(C)}}$ & 0.78 (0.11) & 0.64 (0.07) & 0.86 (0.02) \\
$\ell_\text{MSE}^{\text{(C,O)}}$ & 0.73 (0.09) & 0.69 (0.06) & 0.68 (0.11) \\
$\ell_\text{MSE}^{\text{(C,S)}}$ & 0.70 (0.10) & 0.62 (0.05) & 0.77 (0.02) \\
$\ell_\text{MSE}^{\text{(O,S)}}$ & 0.70 (0.09) & \textbf{0.55 (0.06)} & 0.58 (0.03) \\
$\ell_\text{MSE}^{\text{(C,O,S)}}$ & 0.65 (0.09) & \textbf{0.55 (0.05)} & 0.51 (0.02) \\
$\ell_\text{MAE}$ & 0.77 (0.09) & 0.78 (0.11) & 0.71 (0.02) \\
$\ell_\text{MAE}^{\text{(C)}}$ & 0.72 (0.10) & 0.72 (0.09) & 0.65 (0.03) \\
$\ell_\text{MAE}^{\text{(O)}}$ & 0.77 (0.08) & 0.69 (0.06) & 0.77 (0.14) \\
$\ell_\text{MAE}^{\text{(C)}}$ & 0.78 (0.10) & 0.64 (0.07) & 0.86 (0.05) \\
$\ell_\text{MAE}^{\text{(C,O)}}$ & 0.70 (0.08) & 0.67 (0.05) & 0.54 (0.08) \\
$\ell_\text{MAE}^{\text{(C,S)}}$ & 0.69 (0.11) & 0.67 (0.06) & 0.83 (0.04) \\
$\ell_\text{MAE}^{\text{(O,S)}}$ & 0.70 (0.09) & 0.62 (0.06) & 0.34 (0.03) \\
$\ell_\text{MAE}^{\text{(C,O,S)}}$ & 0.64 (0.09) & 0.61 (0.06) & 0.33 (0.02) \\
$\ell_\text{MSE}^{(\text{O}_\text{S})}$ & 0.94 (0.20) & 0.72 (0.07) & 0.30 (0.03) \\
$\ell_\text{MSE}^{(\text{C},\text{O}_\text{S})}$ & 0.93 (0.19) & 0.73 (0.09) & 0.30 (0.03) \\
$\ell_\text{MSE}^{(\text{S},\text{O}_\text{S})}$ & 0.74 (0.10) & 0.59 (0.05) & 0.32 (0.03) \\
$\ell_\text{MSE}^{(\text{C}, \text{O}_\text{S}, \text{S})}$ & 0.75 (0.10) & 0.59 (0.06) & 0.31 (0.02) \\
$\ell_\text{MAE}^{(\text{O}_\text{S})}$ & 0.94 (0.23) & 0.70 (0.07) & 0.33 (0.03) \\
$\ell_\text{MAE}^{(\text{C},\text{O}_\text{S})}$ & 0.97 (0.27) & 0.69 (0.07) & 0.33 (0.03) \\
$\ell_\text{MAE}^{(\text{S},\text{O}_\text{S})}$ & 0.81 (0.12) & 0.62 (0.07) & 0.34 (0.04) \\
$\ell_\text{MAE}^{(\text{C}, \text{O}_\text{S}, \text{S})}$ & 0.79 (0.11) & 0.61 (0.07) & 0.34 (0.04) \\
$\ell_\text{SPO+}$ & 0.61 (0.13) & 0.57 (0.06) & 0.29 (0.03) \\
$\ell_\text{PFYL}$ & \textbf{0.60 (0.07)} & 0.57 (0.06) & \textbf{0.26 (0.03)} \\
\bottomrule
\end{tabular}
\endgroup
\caption{Test absolute regret normalized on MSE (standard deviation), averaged over 20 seeds. In bold the best performing method.}
\label{tab:bench}
\end{table}

\begin{table*}[t]
\centering
\begingroup\footnotesize
\setlength{\tabcolsep}{1pt}
\renewcommand{\arraystretch}{1.2}
\begin{tabular}{lcccccccccccccccccccccc}
\toprule
 & \multicolumn{2}{c}{CVRP20} & \multicolumn{2}{c}{CVRP30} & \multicolumn{2}{c}{CVRP40} & \multicolumn{2}{c}{TSP100} & \multicolumn{2}{c}{TSP250} & \multicolumn{2}{c}{TSP500} & \multicolumn{2}{c}{TSP1000} & \multicolumn{2}{c}{CAKS100} & \multicolumn{2}{c}{CAKS500} & \multicolumn{2}{c}{CAKS1000} & \multicolumn{2}{c}{KS500} \\
 & regret & time & regret & time & regret & time & regret & time & regret & time & regret & time & regret & time & regret & time & regret & time & regret & time & regret & time \\
\midrule
$\ell_\text{MSE}$ & 1.00 & 20 & 1.00 & 39 & 1.00 & 10 & 1.00 & 39 & 1.00 & 30 & 1.00 & 28 & \textbf{1.00} & \textbf{108} & 1.00 & 14 & 1.00 & 31 & 1.00 & 112 & 1.00 & 15 \\
$\ell_\text{MSE}^{(\text{O}_\text{S})}$ & 1.76 & 27 & 0.92 & 35 & 0.85 & 24 & \textbf{0.27} & \textbf{155} & 0.26 & 881 & 0.46 & 1798 & 7.70 & 2774 & 1.06 & 22 & 1.17 & 31 & \textbf{1.18} & \textbf{46} & 0.71 & 33 \\
$\ell_\text{MSE}^{\text{(O,S)}}$ & 0.62 & 21 & 0.58 & 41 & 0.58 & 12 & 0.61 & 23 & 0.75 & 44 & 1.54 & 60 & 3.79 & 270 & 0.97 & 15 & \textbf{0.96} & \textbf{27} & \textbf{0.95} & \textbf{96} & \textbf{0.55} & \textbf{17} \\
$\ell_\text{MSE}^{\text{(C,O,S)}}$ & 0.54 & 122 & 0.56 & 1196 & \multicolumn{2}{c}{*} & 0.60 & 444 & 0.72 & 3579 & \multicolumn{2}{c}{*} & \multicolumn{2}{c}{*} & 0.97 & 65 & 0.95 & 206 & 0.95 & 358 & 0.56 & 130 \\
$\ell_\text{MAE}^{\text{(O,S)}}$ & \textbf{0.41} & \textbf{21} & \textbf{0.34} & \textbf{40} & \textbf{0.31} & \textbf{12} & \textbf{0.28} & \textbf{22} & \textbf{0.31} & \textbf{43} & \textbf{0.36} & \textbf{56} & \textbf{0.58} & \textbf{280} & \textbf{0.84} & \textbf{16} & 0.98 & 31 & 0.98 & 94 & 0.62 & 16 \\
$\ell_\text{MAE}^{\text{(C,O,S)}}$ & 0.41 & 140 & \textbf{0.33} & \textbf{1119} & \multicolumn{2}{c}{*} & 0.27 & 457 & 0.30 & 3691 & \multicolumn{2}{c}{*} & \multicolumn{2}{c}{*} & 0.83 & 54 & 0.98 & 187 & 0.98 & 356 & 0.64 & 124 \\
$\ell_\text{SPO+}^{lr=0.1}$ & 0.46 & 2433 & 0.35 & 3628 & \textbf{0.29} & \textbf{3633} & \textbf{0.12} & \textbf{1801} & \textbf{0.10} & \textbf{3641} & \textbf{0.14} & \textbf{3718} & \textbf{0.31} & \textbf{3790} & 0.82 & 560 & 0.76 & 735 & 0.77 & 1021 & 0.64 & 667 \\
$\ell_\text{SPO+}$ & 0.40 & 3604 & \textbf{0.31} & \textbf{3642} & 0.30 & 3634 & \textbf{0.11} & \textbf{3603} & 0.15 & 3620 & 0.34 & 3690 & 0.83 & 4442 & 0.81 & 1202 & 0.76 & 2279 & 0.77 & 2607 & 0.57 & 3602 \\
$\ell_\text{SPO+}^{(\text{relax})}$ & 0.54 & 241 & 0.38 & 487 & 0.32 & 443 & 0.13 & 3600 & \textbf{0.11} & \textbf{3604} & \textbf{0.18} & \textbf{3628} & \textbf{0.46} & \textbf{3699} & \textbf{0.81} & \textbf{53} & \textbf{0.76} & \textbf{135} & \textbf{0.77} & \textbf{294} & 0.57 & 101 \\
$\ell_\text{SPO+}^{(\text{cache }5\%)}$ & 0.41 & 735 & 0.45 & 512 & 0.65 & 488 & 0.50 & 431 & \textbf{0.24} & \textbf{664} & 0.79 & 589 & 1.00 & 1348 & 0.82 & 247 & \textbf{0.75} & \textbf{316} & 0.78 & 345 & 0.60 & 311 \\
$\ell_\text{SPO+}^{(\text{cache }0\%)}$ & 0.79 & 278 & 1.04 & 298 & 0.86 & 129 & 0.81 & 447 & 0.75 & 494 & 0.94 & 323 & 0.98 & 266 & 0.96 & 164 & 1.01 & 198 & 0.99 & 299 & 0.86 & 259 \\
$\ell_\text{MAP}^{(\text{cache }5\%)}$ & 0.56 & 567 & 0.45 & 944 & 0.45 & 906 & \textbf{0.22} & \textbf{463} & \textbf{0.17} & \textbf{967} & 0.81 & 548 & 1.01 & 947 & 1.03 & 140 & 0.96 & 314 & 1.17 & 346 & 0.94 & 309 \\
$\ell_\text{MAP}^{(\text{cache }0\%)}$ & 1.50 & 57 & 2.06 & 81 & 1.09 & 32 & 1.07 & 77 & 1.06 & 188 & 0.97 & 182 & 1.00 & 212 & 0.98 & 42 & 1.00 & 84 & 1.00 & 158 & 1.14 & 65 \\
$\ell_\text{CaVE}$ & \textbf{0.35} & \textbf{1043} & \multicolumn{2}{c}{*} & \multicolumn{2}{c}{*} & 1.00 & 3716 & \multicolumn{2}{c}{*} & \multicolumn{2}{c}{*} & \multicolumn{2}{c}{*} & 4.27 & 417 & 3.14 & 629 & 1.99 & 913 & 1.83 & 479 \\
$\ell_\text{AVA}$ & \multicolumn{2}{c}{*} & \multicolumn{2}{c}{*} & \multicolumn{2}{c}{*} & \multicolumn{2}{c}{*} & \multicolumn{2}{c}{*} & \multicolumn{2}{c}{*} & \multicolumn{2}{c}{*} & 0.96 & 115 & 0.91 & 330 & 0.93 & 719 & 0.88 & 295 \\
\midrule
 & \multicolumn{2}{c}{CVRP20} & \multicolumn{2}{c}{CVRP30} & \multicolumn{2}{c}{CVRP40} & \multicolumn{2}{c}{TSP100} & \multicolumn{2}{c}{TSP250} & \multicolumn{2}{c}{TSP500} & \multicolumn{2}{c}{TSP1000} & \multicolumn{2}{c}{CAKS100} & \multicolumn{2}{c}{CAKS500} & \multicolumn{2}{c}{CAKS1000} & \multicolumn{2}{c}{KS500} \\
\midrule
train & \multicolumn{2}{c}{900} & \multicolumn{2}{c}{900} & \multicolumn{2}{c}{450} & \multicolumn{2}{c}{900} & \multicolumn{2}{c}{900} & \multicolumn{2}{c}{450} & \multicolumn{2}{c}{200} & \multicolumn{2}{c}{900} & \multicolumn{2}{c}{900} & \multicolumn{2}{c}{900} & \multicolumn{2}{c}{900} \\
val & \multicolumn{2}{c}{100} & \multicolumn{2}{c}{100} & \multicolumn{2}{c}{50} & \multicolumn{2}{c}{100} & \multicolumn{2}{c}{100} & \multicolumn{2}{c}{50} & \multicolumn{2}{c}{50} & \multicolumn{2}{c}{100} & \multicolumn{2}{c}{100} & \multicolumn{2}{c}{100} & \multicolumn{2}{c}{100} \\
test & \multicolumn{2}{c}{500} & \multicolumn{2}{c}{500} & \multicolumn{2}{c}{250} & \multicolumn{2}{c}{500} & \multicolumn{2}{c}{500} & \multicolumn{2}{c}{250} & \multicolumn{2}{c}{100} & \multicolumn{2}{c}{500} & \multicolumn{2}{c}{500} & \multicolumn{2}{c}{500} & \multicolumn{2}{c}{500} \\
\bottomrule
\end{tabular}
\endgroup
\caption{All results for the second set of experiments, including a summary of used train sizes. Regret is test absolute regret normalized on the MSE regret averaged over 3 seeds, time is in seconds. Bold denotes Pareto-optimal methods, i.e. no other method has lower-or-equal regret and runtime (within 30s considered equal).  $*$ denotes cases where a single solve over all training instances took more then 3600s.}
\label{tab:scale}
\end{table*}

\end{document}